\documentclass[conference]{IEEEtran}
\usepackage{fancyhdr}
\usepackage{float}                                                          
\usepackage{cite}
\usepackage{graphicx}
\usepackage{epstopdf}
\usepackage{amsfonts}

\usepackage{amsmath,amssymb,amsthm} 
\theoremstyle{definition}
\usepackage{algorithm}
\usepackage{algorithmic}
\usepackage[small]{caption}
\usepackage{subcaption}
\usepackage{color}
\usepackage{bm}
\usepackage[utf8]{inputenc}
\usepackage[OT1]{fontenc}
\usepackage[]{units}
\usepackage{blkarray}

\graphicspath{{.}{figures/}}

\newcommand{\ep}{\theta}

\newcommand{\dr}{\dot{r}}

\newcommand{\bH}{\mathcal{H}}

\newcommand{\bL}{\mathcal{L}}
\newcommand{\bN}{\mathcal{N}}

\newcommand{\bG}{\mathcal{G}}
\newcommand{\bE}{\mathcal{E}}

\newcommand{\bS}{\mathcal{S}}

\newcommand{\RR}{{\mathbb R}}

\newcommand{\bx}{\bm{x}}
\newcommand{\bX}{\bm{X}}
\newcommand{\bu}{\bm{u}}
\newcommand{\bU}{\bm{U}}

\newcommand{\bJ}{\mathcal{J}}

\DeclareGraphicsExtensions{.pdf,.png,.eps,.ps, .jpg}
\newtheorem{theorem}{Theorem}

\newtheorem{definition}{Definition}

\newtheorem{example}{Example}

\ifCLASSINFOpdf
\else
\fi
\begin{document}
%
\title{A Distributed Model Predictive Control Framework for Road-Following Formation Control of Car-like Vehicles (Extended Version)}

\author{\IEEEauthorblockN{Xiangjun Qian$^{1}$, Florent Altché$^{2,1}$,   Arnaud de La Fortelle$^{1}$, Fabien Moutarde$^{1}$}
\IEEEauthorblockA{$^{1}$PSL - Research University, Centre	for Robotics, 60 Bd	St Michel \\ 75006 Paris, France}
$^{2}$École des Ponts ParisTech, Cité Descartes, 6-8 Av Blaise Pascal\\
 77455 Champs-sur-Marne, France\\
 Email: xiangjun.qian@mines-paristech.fr}


%


\maketitle

\thispagestyle{fancy}
\fancyhead{}
\lhead{}
\lfoot{}
\cfoot{}
\rfoot{}
\renewcommand{\headrulewidth}{0pt}
\renewcommand{\footrulewidth}{0pt}

\begin{abstract}
This work presents a novel framework for the formation control of multiple autonomous ground vehicles in an on-road environment. Unique challenges of this problem lie in 1) the design of collision avoidance strategies with obstacles and with other vehicles in a highly structured environment, 2) dynamic reconfiguration of the formation to handle different task specifications. In this paper, we design a local MPC-based tracking controller for each individual vehicle to follow a reference trajectory while satisfying various constraints (kinematics and dynamics, collision avoidance, \textit{etc.}). The reference trajectory of a vehicle is computed from its leader's trajectory, based on a pre-defined formation tree. We use logic rules to organize the collision avoidance behaviors of member vehicles. Moreover, we propose a methodology to safely reconfigure the formation on-the-fly. The proposed framework has been validated using high-fidelity simulations.
\end{abstract}

\section{Introduction}

Over the past decade, we have witnessed some major developments of autonomous driving technologies. With the vision of mass deployment of autonomous vehicles on the roads, the problem of coordinating groups of autonomous vehicles has gained a lot of attention. Platoons, as a type of on-road coordination, have been extensively studied~\cite{69979,981409,Chan}. Vehicles in a platoon move along a road in a linear formation with small longitudinal gaps. The advantage of platoons lies mainly in a reduced fuel consumption due to lower air drag and an increased road throughput thanks to smaller inter-vehicular distance. 

We consider an extended version of platoons in which multiple vehicles can enter a formation with both longitudinal and lateral separations. We expect that this extension can find its applications in cooperative tasks, for instance protecting a VIP vehicle, snowplowing (see example in~\cite{Ono2015}), cooperative lane change, \textit{etc.}.   

There has been a significant volume of work on generic robot formation control problems~\cite{4123995,consolini2008leader,ren2002virtual,ren2004decentralized, balch1998behavior}. References~\cite{4123995,consolini2008leader} propose a leader-following approach in which a leader follows a reference trajectory, while other robots maintain a relative position offset with the leader. Virtual structure based techniques~\cite{ren2002virtual,ren2004decentralized} consider the formation as a moving rigid structure, where robots locally track their reference trajectories based on the trajectory of the structure. Distributed Model Predictive Control (MPC) is a promising technique for the formation control problem thanks to its generality and reconfigurability. The theoretical foundation of applying distributed MPC on the multi-vehicle coordination problem has been laid in~\cite{dunbar2004distributed,borrelli2005hybrid}. It is shown in~\cite{dunbar2004distributed} that we can stabilize a formation by only formulating local MPC problems and exchanging previously calculated optimal trajectories, provided that the previously calculated trajectories do not deviate too much from the ones being actually followed. In~\cite{borrelli2005hybrid}, a hybrid system approach combining logical rules with MPC is proposed. This approach allows to regulate the interactions between vehicles in an organized and predictable way, thus avoiding unexpected formation behaviors and therefore improving stability and feasibility. 

The literature in generic robot formation control can be a source of inspiration for the on-road formation control problem. However, unique challenges set the on-road formation problem apart from its generic counterpart. First, vehicles must intelligently balance the goal of formation keeping with obstacle/collision avoidance in a highly constrained on-road environment. Second, vehicles must be able to deliberatively reconfigure their formation to handle different task specifications.

Previous work exists on the road-following formation control problem. Reference~\cite{qian2016hierarchical} represents the formation as a virtual structure and uses a hierarchical Model Predictive Control framework to maintain the formation. However, the intra-formation collision avoidance is not considered and the virtual structure is static. In~\cite{arad}, a hybrid two-layer framework is proposed, in which an outer layer working in discrete modes safely coordinates the vehicles to find collision-free trajectories, while an inner layer is tasked with trajectory tracking. Moreover, the formation is determined in advance and cannot be reconfigured.  Other approaches include Laplacian-based control~\cite{7225676}, graph searching methods~\cite{Ono2015}, \textit{etc.}.

This paper proposes and validates a novel on-road formation control framework established on the theoretical results of distributed MPC, upon which we propose multiple adaptations to handle the challenges raised by the on-road driving setting.  We adopt a fully distributed paradigm in which each vehicle is equipped with a MPC tracking controller. The module computes a sequence of control inputs to track a reference trajectory  while satisfying various constraints (kinematics and dynamics, collision avoidance, \textit{etc.}). The reference trajectory of a vehicle is computed from its leader's trajectory, based on a pre-defined formation tree. The major contribution of this paper is the use of hybrid system techniques to organize the collision avoidance behaviors of member vehicles. To the best of our knowledge, it is the first time that such design is proposed in this problem setting. 

The hybrid approach allows to combine the optimization part of the MPC technique with logic rules that relieve the optimization process from lengthy collision constraints computations. Moreover, logic rules can be made semantically meaningful, which could lead to easier interactions between automated vehicles and people (be it other drivers or pedestrians, cyclists, \textit{etc.}). Finally, our framework allows to safely modify the formation on-the-fly. In this article, we present preliminary results validated by simulation, demonstrating this framework is suitable for actual implementation.

The rest of the paper is organized as follows. In Section II, we present the MPC-based trajectory generation for vehicles. Section III presents the modeling of the on-road formation control problem. Section IV shows how this coordination framework can be used for dynamic formation changes. Section V validates our approach through simulations, and Section VI concludes the the paper.

\section{MPC-based Trajectory Generation for Individual Vehicles}
\label{sec:model} 
\subsection{Road-following Coordinate System}
\label{sec:cs}
We first define a road-following coordinate system. We assume that the centerline of the road can be described as a curve $\Gamma \in \RR^2$ with $C^3$ continuity, ensuring that the curvature $c$ of the curve and its derivative $\kappa$ exist. We define a Frenet coordinate system $(s,r)$, where $s$ is the curvilinear abscissa along $\Gamma$, and $r$ the lateral deviation. The left and right boundaries of the road are defined as continuous functions $\bar{r}(s)$ and $\underline{r}(s)$. To ensure the bijection from the $(s,r)$ frame to a Cartesian frame, we require that for all $(s,r)$ such that $\underline{r}(s) \leq r \leq \bar{r}(s)$:
\begin{equation}
1 - r c(s) < 0,
\end{equation}
where  $c (s)$ is the curvature profile of the road's centerline at point $s$.

\subsection{Kinematic Bicycle Model in the Frenet Frame}\label{sec:vm}
We adopt the common practice in vehicle control to set up a two-layer control structure, in which an inner-loop controller stabilizes the system and a outer-loop planning module plans the trajectory according to the goal configuration. If the inner-loop controller is fast enough to track the given reference trajectories, the outer-loop dynamics can be approximately described as a standard kinematic bicycle model~\cite{Kong}.

As we consider on-road autonomous driving, we formulate the kinematic bicycle model in the road-following coordinate system as follows:
\begin{subequations}
\begin{align}
&\dot{s} = v \cos\ep \left(\frac{1}{1 - r c (s)}\right),\label{eq:ds} \\
&\dr = v \sin\ep, \\
&\dot{\ep} = v k - v  \cos\ep \left(\frac{c(s)}{1 - r c (s)}\right) , \\ 
&\dot{v} = a, \\
&\dot{k} = \kappa. \label{eq:dk}
\end{align}
\label{eq:eq}
\end{subequations} 
$c(s)$ is assumed to be known, for instance from maps. Other state variables $r$, $v$, $\ep$ and $k$ are respectively the lateral offset, the vehicle speed, the heading alignment error of the vehicle with regard to the  centerline and the curvature of the vehicle's trajectory. The chosen control inputs are the acceleration $a$ and the first derivative of curvature, $\kappa$. We note $\bx =[s,r,v,\ep,k] \in \bX$ the state of the vehicle, and $\bu = [a, \kappa]\in \bU$ its control. $\bX$ and $\bU$ are compactly written form of the state constraint and the control constraint. We note $\dot{\bx} = f(\bx,\bu)$ the system \eqref{eq:ds}-\eqref{eq:dk}. When there is a need to differentiate between different vehicles, subscripts like $i,j$ can be added to those variables.

More details on the transformation of the kinematic bicycle model from a Cartesian frame to the Frenet frame can be found in~\cite{Weiskircher}.
\begin{figure}[t!] 
	\centering
	\includegraphics[width=0.6\linewidth]{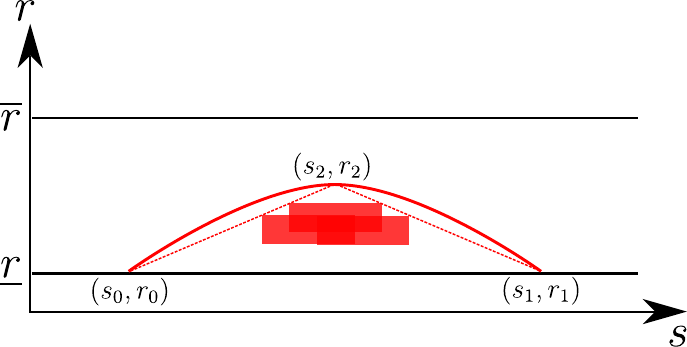}
	\caption{Approximation of an obstacle region by a parabola. }
	\label{fig:obs}
\end{figure}
\subsection{Obstacle Modeling}\label{sec:obmodel}
Vehicle trajectories are generated using a receding horizon scheme based on optimization techniques. However, the convergence to a global optimum cannot be guaranteed. With one obstacle in the middle of the road, there are two homotopy classes of trajectories, respectively corresponding to avoiding the obstacle from the left and from the right. This situation leads to multiple stationary points for the optimization problem. In this paper, we heuristically address the problem by affecting obstacles to their closest side of the road, thus eliminating one homotopy class. A more detailed discussion on this issue can be found in\cite{bender2015combinatorial}.

Another issue in obstacle modeling is that the optimization-based scheme requires obstacle boundaries to be differentiable so that traditional gradient-based methods can be deployed; in general, road obstacles do not display such properties. To overcome this issue, we approximate obstacles by bounding triangles, and we use the unique parabola going through the three points of the triangle to get smooth obstacle boundaries, as represented in Fig.~\ref{fig:obs}. Note that the parameters of this parabola can be found quickly by solving a system of three linear equations. In what follows, we let $\mathcal O$ be the set of all obstacles and, for an obstacle $o \in \mathcal O$ we define $h_o(\bx)$ such that $h_o(\bx) \leq 0$ if, and only if, the vehicle is outside the concavity of the parabola corresponding to $o$ at time $t$.

\subsection{MPC formulation}\label{sec:mpc}   
The control objective for each vehicle is to track its reference trajectory. Therefore, we consider the following cost function that penalizes the aggregated deviation from the reference trajectory over the planning horizon. More specifically, the cost function is given as:
\begin{eqnarray}
\bJ (\bx , \bu ) &=& ||\bx(T) -\bx_{ref}(T)||_{P}^2 \label{eq:obj}\\
&&+\int_{t_0}^{t_0 + T}\left( ||\bx(t) -\bx_{ref}(t)||_{Q}^2 + ||\bu(t)||_{R}^2 \right)\nonumber
\end{eqnarray}
where $t_0$ is the current time instant and $T$ is the prediction horizon. $P$, $Q$ and $R$ are positive diagonal weighting matrices of respective dimensions 5, 5 and 2. The running cost term $\int_{t_0}^{t_0 + T}( ||\cdot||_{Q}^2 + ||\cdot||_{R}^2 )$ penalizes the deviation from the desired trajectory, as well as the control effort. The terminal cost term $||\cdot||_{P}^2$ is set to 0 in this paper.  We will further discuss the construction of the reference trajectory $\bx_{ref}$ in the following section, as it is the major enabler of our formation control framework. 

At each sampling time $t = t_0 $, we solve the following constrained optimization problem:
\begin{equation}\label{eq:mpc0}
\hspace{-2cm}\min_{\bu } \bJ (\bx , \bu ), 
\end{equation} 
subj. to $\forall t \in [t_0, t_0 +T],$
\begin{subequations}
\begin{align}
&\dot{\bx}(t) = f(\bx(t),\bu(t)),\label{eq:cons} \\
&\bx(t) \in \bX\qquad\textrm{and}\qquad \bu(t) \in \bU , \label{eq:bound} \\
&  -\bar{a}_{i,lat} \leq v(t)^2 k(t) \leq \bar{a}_{i,lat} , \label{eq:lat} \\ 
& h_o(\bx(t)) \leq 0, \quad\forall o \in \mathcal{O} \label{eq:obs}.
\end{align}
\label{eq:mpc}
\end{subequations}
Constraint~\eqref{eq:cons} is the compactly-written form of the vehicle model. Eq.~\eqref{eq:bound} captures various bound constraints on speed, lateral deviation, acceleration, \textit{etc}. Eq.~\eqref{eq:lat} limits the lateral acceleration of the vehicle and Eq.~\eqref{eq:obs} ensures the obstacle avoidance constraints for all nearby obstacles, using the method described in Section~\ref{sec:obmodel}.

The resulting optimal trajectory is then (periodically) fed to a low level controller for tracking.

\begin{figure}[t!] 
\centering
\includegraphics[width=0.5\linewidth]{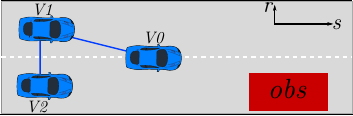}
\caption{A three vehicle formation with an obstacle on the road.}
\label{fig:3v}
\end{figure}

\section{Cooperative Formation control}
This section discusses the formation-control scheme that allows vehicles to maintain a convoy formation while avoiding obstacles and other vehicles. 
\subsection{Formation Model}
\label{sec:cm}
Consider a set $\bN = \{0,..,N\}$ of $N+1$ vehicles, in which we arbitrarily use $0$ as leader; the leader can either be a physical vehicle, or a virtual reference point.  We define the target shape of the formation through a $ (N + 1) \times 2$ matrix $\bS$ in which each row vector $ (s^d_i,r^d_i)$ encodes the target position of vehicle $i$ relatively to the leader.  To describe the interdependence relations between vehicles in the formation, we define a formation tree as a directed tree $\bG=(\bN,\bE)$, whose nodes are the vehicles of $\bN$ and with a set of edges $\bE$ such that the root node is the leader $0$. Such a tree can be represented as an adjacency matrix $(g_{ij})$ of size $(N+1) \times (N+1)$, in which element $g_{ij}$ equals $1$ if the edge $i \rightarrow j$ is in $\bE$, and $0$ otherwise. Remark that we expect $g_{ij} = 1 \Rightarrow s^d_i \geq s^d_j$ so that a given vehicle needs only take information from the preceding vehicles, although it is not mandatory in our formulation. 
\begin{example}
Consider a triangular formation of three vehicles as shown in Fig.~\ref{fig:3v}. The shape $\bS$ of the formation is defined as
\begin{equation*}
\begin{blockarray}{ccc}
&s^d_i&r^d_i\\
\begin{block}{c(cc)}
0&0&0\\
1&-10&3\\
2&-10&-3\\
\end{block}
\end{blockarray}
\end{equation*} The formation tree $\bG$ defined by the adjacency matrix $g$ is given as: \begin{equation*}
\begin{blockarray}{cccc}
&0&1&2\\
\begin{block}{c(ccc)}
0&0&0&0\\
1&1&0&0\\
2&0&1&0\\
\end{block}
\end{blockarray}
\end{equation*} such that vehicle~1 computes its formation control trajectory relative to vehicle~0, and vehicle~2 computes its trajectory relative to vehicle~1. 
\end{example}

\subsection{Formation Control Scheme}\label{sec:ov}

Vehicles in the formation are assumed to use the MPC-based trajectory generation scheme described in Section~\ref{sec:model}. We propose to adjust the reference trajectories of vehicles to drive them into the desired formation.

Consider an arbitrary pair of vehicles $(i,j)$ such that $g_{ij} =1$. A communication link can be established between $i,j$ such that vehicle $j$ periodically receives information on the planned trajectories of vehicle $i$. Assume that at time $t$ the most up-to-date trajectory for vehicle $i$ received by vehicle $j$ is $\bx_i([t_1,t_1+T_p])$, with $t_1 < t$: the trajectory $\bx_i([t,t + T_p])$ can be simply estimated by simulating the trajectory from $t_1 + T_p$ to $t + T_p$ using the last value of the control. Under the assumption that the communication delay is small, we expect the estimated trajectory $\bx_i([t,t + T_p])$ to remain close to the actually planned trajectory of vehicle $i$. 

The desired relative position of $j$ from $i$ can then be calculated as
\begin{subequations}
\begin{align}
&s^d_{ji} = s^d_j - s^d_i,\\
&r^d_{ji} = r^d_j - r^d_i.
\end{align}
\label{eq:des}
\end{subequations}

Finally, the reference trajectory $\bx_{j,ref}$ can be obtained by offsetting the position components of $\bx_i([t,t+T_p])$ with $s^d_{ij}$ and $r^d_{ij}$ and setting other components to zero.

The proposed approach allows a distributed implementation of the formation control scheme. Only a small amount of information related to the formation definition is synchronized among vehicles, while each vehicle uses local information to plan its own trajectory. 

\subsection{Intra-formation Collision Avoidance}\label{sec:inavoid}

Our aim is to design an intra-formation collision avoidance strategy such that vehicles will not collide in normal on-road driving situations, nor in situations where the formation is perturbed by obstacles. 
\begin{figure}[ht!] 
\centering
\includegraphics[width=0.6\linewidth]{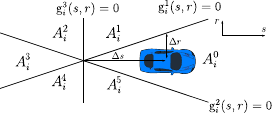}
\caption{Road space partitioning with respect to vehicle $i$.}
\label{fig:ca}
\end{figure}
Remark that the exact rectangle-shaped safety region of a vehicle is non-differentiable, and multiple choices exist for collision avoidance. Taking a 2-vehicle scenario as an example (vehicle~1 and 2 in Fig.~\ref{fig:3v}), if vehicle~1 approaches vehicle~2, vehicle~2 has four strategies to avoid collision: acceleration to the front of vehicle~1, deceleration behind vehicle~1, moving further to the right or circumvent vehicle~1 from its left. The problem becomes exponentially complex as the number of vehicles increases due to its combinatorial nature. In~\cite{borrelli2005hybrid}, a similar problem is handled through the formulation of a mixed integer problem, in which the avoidance decisions are modeled as binary variables. However, the real-time requirement of our algorithm stops us from adopting the same method. 

Here, we adopt a hybrid modeling approach. Let us define an ordered priority list $\bL$ as a permutation of $\{0, 1,...,N\}$, representing the relative priorities between vehicles. With a slight abuse of notation, we let $\bL(i)$ be the rank of $i$ in $\bL$, \textit{i.e.} $\bL(i) = j$ if, and only if, $i$ is the $j$-th element of $\bL$. We enforce the following constraint for any pair of vehicles $i,j \in \bN$:

\medskip
\textbf{Design constraint:} $\bL(i) < \bL(j) \iff s^d_i \geq s^d_j$
\medskip

Under this constraint, the list $\bL$ encodes a traffic rule for the formation of vehicles: each vehicle is only responsible for avoiding collisions with the higher-priority vehicles on the road, \textit{i.e} with vehicles preceding it in $\bL$.

To account for the existence of different homotopy classes, we divide the space outside the collision region in six areas: for a vehicle $i$, we partition the space using three affine functions $g^1_i(s,r) = 0$, $g^2_i(s,r) = 0$ and $g^3_i(s,r) = 0$ as shown in Fig.~\ref{fig:ca}:
\begin{subequations}
\begin{align}
&g^1_{i}(s,r) = -\frac{r - r_i}{\Delta r} + \frac{s - s_i}{\Delta s} + 1, \\
&g^2_{i}(s,r) = \frac{r - r_i}{\Delta r} + \frac{s - s_i}{\Delta s} - 1, \\
&g^3_{i}(s,r) = \frac{s - s_i}{\Delta s} + 1. 
\end{align}
\label{eq:part}
\end{subequations} 
where $\Delta s$ and $\Delta r$ are geometric parameters described in Fig.~\ref{fig:ca}. For each vehicle $i$, these affine functions define a partition of the $(s,r)$ plane in six sub-spaces labeled as $A^m_i, m\in\{0..5\}$ as shown in Fig.~\ref{fig:ca}. By definition, vehicle $i$ is always inside region $A^0_i$, and we enforce safety by preventing vehicles with lower priority than $i$ from entering $A^0_i$. In what follows, we call $A^0_i$ the \textit{protected region} for $i$, and the union of subsets $A^m_i, m\in\{1\dots 5\}$ forms the \textit{safe region} with respect to $i$.  

For an arbitrary vehicle $j$, the following logic rule is introduced for all $i \in \bN$:
\begin{equation}
\bL(i) < \bL(j) \Rightarrow (s_j,r_j) \in \bigcup_{m = \{1..5\}} A^m_i.
\label{eq:sr}
\end{equation} 

Rule~\eqref{eq:sr} effectively forces each vehicle to remain in the safe region with respect to the vehicles having higher priority in $\bL$. In a classic MPC setting, a decision variable would be required to select in which of the subspaces $A^m_i, m\in\{1\dots 5\}$ vehicle $j$ should be. In this article we adopt a different, simpler approach consisting of using the shape matrix $\bS$ to implicitly constrain the choice of the homotopy class as follows: $\forall i \in \bN,$
\begin{subequations}
\begin{align}
\bL(i) < \bL(j) \wedge s^d_{ji} \geq - \Delta s \wedge r^d_{ji} > 0  &\Rightarrow g^1_i(s_j,r_j) \leq 0, \\
\bL(i) < \bL(j) \wedge s^d_{ji} \geq - \Delta s \wedge r^d_{ji} < 0  &\Rightarrow g^2_i(s_j,r_j) \leq 0, \\
\bL(i) < \bL(j) \wedge s^d_{ji} <  - \Delta s   &\Rightarrow g^3_i(s_j,r_j) \leq 0. 
\end{align}
\label{eq:posde_cons}
\end{subequations} 
Note that, although the space represented by $\bigcup_{m = \{1\dots 5\}} A^m_i$ is non-differentiable, the above constraints are linear (and therefore differentiable). As a trade-off, the equations in~\eqref{eq:posde_cons} are more restrictive than rule~\eqref{eq:sr}. Constraints~\eqref{eq:posde_cons} are then incorporated into the MPC formulation~\eqref{eq:mpc0}-\eqref{eq:mpc} for enforcement. Also note that, in the above constraints, the shape matrix and value of $\Delta s$ are parameters and not variables; therefore, only one of the three above constraint is actually added to the problem at each step.

\begin{example}
Let $\bL = \{0,1,2\}$ be the priority list for the formation illustrated in Fig.~\ref{fig:3v}. The constraint for vehicle~1 can be computed from~\eqref{eq:posde_cons} as $g^3_0(s_1,r_1) \leq 0$ such that the safe region (relative to vehicle 0) is $A^2_0 \cup A^3_0 \cup A^4_0$, forcing vehicle~1 to stay behind vehicle 0. The constraint for vehicle~2 is $g^3_0(s_2,r_2) \leq 0 \wedge g^2_1(s_2,r_2) \leq 0$, such that vehicle~2's safe region is $(A^2_0 \cup A^3_0 \cup A^4_0)\cap (A^3_1\cup A^4_1 \cup A^5_1)$: vehicle~2 must stay behind vehicle 0, and stay on the right-hand or behind vehicle~1. 
\end{example}

\subsection{Discussions on stability and feasibility}\label{sec:stab}
Here we briefly discuss the stability and feasibility issues of the proposed framework.

This framework is an application and implementation of the more theoretical result of~\cite{dunbar2004distributed}, where a general distributed model predictive control of a dynamically decoupled multi-agent system is considered. It is shown that  stability of the formation can be ensured if, for each agent, the difference between trajectories resulting from two consecutive MPC optimizations is smaller than $\delta^2 q$ (which is called the \textit{compatibility constraint}), where $\delta$ is the update interval and $q$ is a properly chosen constant. The philosophy is that, provided each agent remains sufficiently close to the trajectory other agents base their own trajectory on, the global system remains stable.

It is also shown in their simulation that even without the constraint, the formation is stable thanks to the inherently consistent nature of MPC (although the theoretical guarantee is lost). On the other hand, adding compatibility constraints may introduce sluggish behaviors as agents cannot react quickly to unexpected changes in the environment. One condition that requires the compatibility constraint is when multiple maneuver choices (local optima) exist in the MPC formulation. In this case, the trajectories obtained from successive optimizations can be largely different as the solver may switch between these different local optima.

In our framework, we opt to not use the compatibility constraint. The hybrid modeling of intra-formation collision avoidance~(\ref{eq:posde_cons}) suppresses the possibility of multiple local optima because the rules define a unique homotopy class of trajectories. The resulting formation is experimentally shown to be stable.

Another issue is the feasibility of local MPC during iterations. Consider the case of vehicle~1 in Fig.~\ref{fig:3v} suddenly turning right towards vehicle~2; the collision constraint can be violated due to the maneuver limits of vehicle~2 and the delay in communication. A possible solution is to use the invariant set theory~\cite{keviczky2008decentralized} so that vehicle~2 is required to guard against all possible maneuvers of vehicle~1 during the MPC iteration. However, this method makes the MPC more conservative and requires more computational power. Here, we choose to design the vehicle formation to be sufficiently loose and transform the safety constraint (using $g^1_i$ as an example) to be a soft constraint as
\begin{equation}
g^1_i(s_j,r_j) \leq \varepsilon, \qquad\varepsilon \geq 0.
\end{equation}  
The occasional constraint violation is therefore permitted, but will be penalized by a quadratic term $K\varepsilon^2$ added to \eqref{eq:mpc0}.    

\section{Dynamic Formation Reconfiguration}
Let $\bG$ be a formation tree, $\bS$ a shape and $\bL$ a priority list, all compatible: we denote by $\bH = (\bG, \bS, \bL)$ the corresponding formation. We give the following definition:
\begin{definition}[Isormorphic formation]
Two formations $\bH_1$, $\bH_2$ are said to be isomorphic if $\bL_1 = \bL_2$. 
\end{definition}
In this paper, we only consider isomorphic formation changes: the reconfiguration of $\bS$ and/or $\bG$. Non-isomorphic formation changes that involve the modification of $\bL$ are a topic for future research.

The reconfiguration of $\bG$ only affects the behaviors of vehicles when the formation is perturbed. Thus we are free to modify $\bG$ as long as the tree structure covers all nodes. On the other hand, we cannot arbitrarily modify the shape matrix $\bS$ due to the design of intra-formation collision avoidance constraint~\eqref{eq:posde_cons}. Consider the case of two vehicles $i,j$ with $\bL(i) < \bL(j)$, where vehicle $j$ is in the partition $A^5_i$ of vehicle $i$. Assuming that we reconfigure the shape matrix, if the new goal configuration of vehicle $j$ resides in $A^ 4_i$, then vehicle $j$ can plan a trajectory to its goal configuration because both $A^ 4_i$ and $A^ 5_i$ belong to the constraint $g^2_i(s,r) \leq 0$. However, if the goal configuration is in $A^2_i$, vehicle $i$ cannot plan a feasible trajectory under the current formulation as $A^2_i$ is not a feasible region under the current constraint $g^2_1$. It is necessary to set an intermediate goal configuration at $A^3_i$. Once vehicle $j$ reaches $A^3_i$, the collision avoidance constraint must be switched from $g^2_i$ to $g^3_i$ and then vehicle $j$ can continue to move towards its goal configuration. This issue is a side effect of the space partitioning technique.

We propose the following definitions:
  
\begin{definition}[1-step Reachable Element]
Consider two elements of the partition $A^m_i$ and $A^n_i$; we say $A^n_i$ is 1-step reachable from $A^m_i$ if there exists $l \in \{1,2,3\}$ such that $g^l_i\le 0$ over $A^n_i \cup A^m_i$. 
\end{definition}
This means we need at least one of the rules~(\ref{eq:posde_cons}) to be kept during reconfiguration. We can see, $A^1_i$ is reachable from $A^2_i$ and $A^3_i$; $A^2_i$ is reachable from $A^1_i$, $A^3_i$ and $A^4_i$; $A^3_i$ is reachable from all elements except $A^0_i$; and symmetrically for the other areas.

\begin{definition}[1-step Reachable Formation]
Consider two formations $\bH_1$ and $\bH_2$. We say that $\bH_2$ is 1-step reachable from $\bH_1$  if $\forall i,j \in \bN \text{ s.t. } \bL(i) <\bL(j)$, the configuration of vehicle $j$ with respect to vehicle $i$ in $\bH_2$ is 1-step reachable from $\bH_1$. 
\end{definition}

\begin{theorem}
Consider an arbitrary pair of isomorphic formations $\bH_1$, $\bH_2$, we can transform from $\bH_1$ to $\bH_2$ through a finite sequence of 1-step reachable formations. 
\end{theorem}
The proof is intuitive, since $A^3_i$ is reachable from all elements except $A^0_i$. Therefore, any formation can be transformed to a linear formation in a finite number of steps.

With the above theoretical result, in order to reconfigure the formation to the desired one, we only need to design an intermediate sequence of 1-step reachable formations. Then we can set up a discrete supervisor to control the transformation. 

\begin{figure*}[ht!] 
\centering
\includegraphics[width=\linewidth]{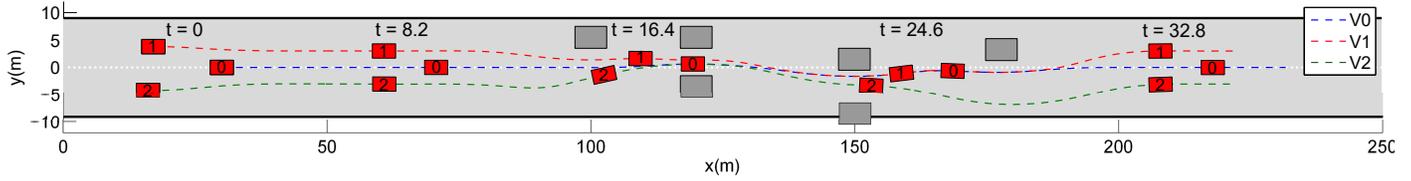}
\caption{First scenario: trajectories of three vehicles.}
\label{fig:exp1traj}
\end{figure*}
\begin{figure*}[ht!] 
    \centering
    \begin{subfigure}[b]{0.32\linewidth}
        \includegraphics[width=\linewidth,height=3.cm]{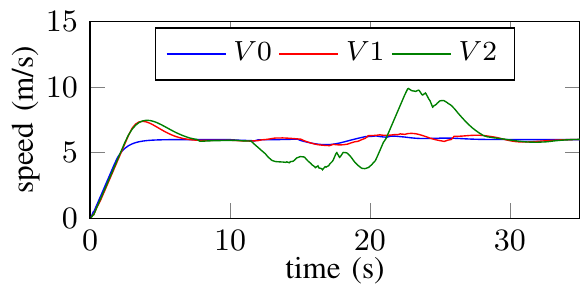}
        \caption{}
        \label{fig:t1}
    \end{subfigure}
    \begin{subfigure}[b]{0.32\linewidth}
        \includegraphics[width=\linewidth,height=3.cm]{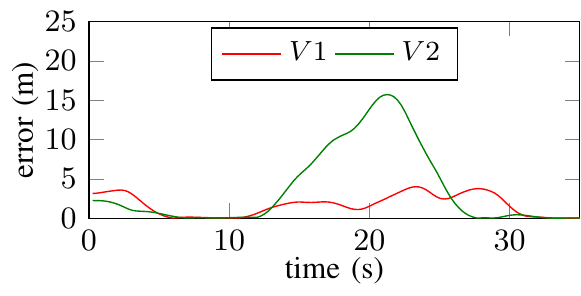}
        \caption{}
        \label{fig:err1}
    \end{subfigure}
    \begin{subfigure}[b]{0.32\linewidth}
        \includegraphics[width=\linewidth,height=3.cm]{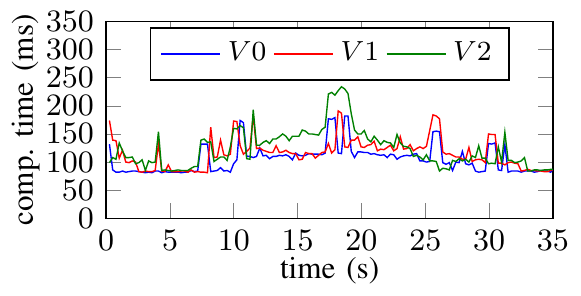}
        \caption{}
        \label{fig:comp1}
    \end{subfigure}
    \caption{First scenario: (a) vehicle speeds, (b) formation error,  (c) computation time.}\label{fig:exp1}
\end{figure*}
\begin{figure}[ht!] 
\centering
\includegraphics[width=\linewidth]{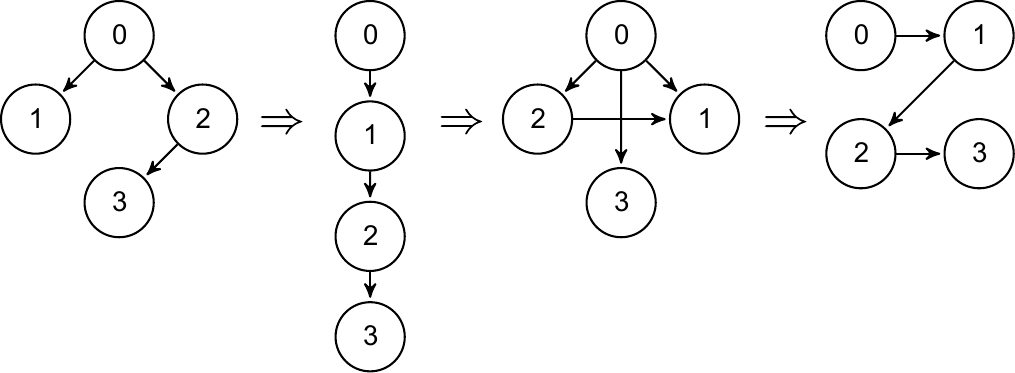}
\caption{A sequence of 1-step reachable isomorphic transformations.}
\label{fig:config}
\end{figure}
\begin{example}\label{ex3}
We consider a formation of four vehicles. Fig.~\ref{fig:config} illustrates a sequence of 1-step reachable isomorphic transformations. The corresponding shape matrices are given as follows.
\begin{center}
\begin{tabular}{c c c c}
$\bS_1$ & $\bS_2$ & $\bS_3$ & $\bS_4$ \\
\begin{blockarray}{cc}
\begin{block}{(cc)}
0&0\\
-10&3\\
-10&-3\\
-20&0\\
\end{block}
\end{blockarray} & 
\begin{blockarray}{cc}
\begin{block}{(cc)}
0&0\\
-10&0\\
-20&0\\
-30&0\\
\end{block}
\end{blockarray} &
\begin{blockarray}{cc}
\begin{block}{(cc)}
0&0\\
-10&-3\\
-10&3\\
-20&0\\
\end{block}
\end{blockarray} &
\begin{blockarray}{cc}
\begin{block}{(cc)}
0&3\\
0&-3\\
-10&3\\
-10&-3\\
\end{block}
\end{blockarray}
\end{tabular}
\end{center}
\end{example}

\section{Experiments and Results}

We have implemented our framework in the high-fidelity robotic simulator Webots~\cite{michel2004webots}. Webots implements a realistic vehicle model with steering dynamic response and non-linear tire models. The proposed algorithm is implemented in C++ and we use the ACADO toolkit~\cite{Houska2011a} to solve the MPC problem.

We have designed two test scenarios. In both scenarios, vehicles are equipped with noise-free localization systems and delay-free communication devices. The desired speed of the formation for both scenarios is given as $v_\bH = \unit[6]{m/s}$.  Vehicle parameters are given as: $0 \leq v_i \leq \unit[10]{m/s}$, $ |a_i| \leq \unit[2.5]{m/s^2}$, $|k_i|  \leq \unit[0.2]{m^{-1}}$, $|\kappa_i| \leq \unit[0.1]{m^{-1}s^{-1}}$ and  $\bar{a}_{i,lat} = \unit[2.5]{m/s^2}$. The parameters used for the leader are $Q_0 = diag(0,4,2, 20, 20)$, $R_0 = diag(1,200)$. The parameters used for the followers are $Q_i = diag(1,2,0,20, 20)$,  $R_i = diag(1,200)$. The constraint violation penalty parameter described in Section~\ref{sec:stab} is selected as $K = 10000$. The prediction horizon for all vehicles is $T_i =\unit[5]{s}$. The trajectory re-planning interval is  $\Delta T_i = \unit[0.256]{s}$. Thus at time $t_0$, a follower has access to the planned trajectory of its leader at time $t_0 - 0.256$.  The parameters for space partition are given as $\Delta s = \unit[10]{m}$ and $\Delta r = \unit[3]{m}$. To quantify the formation error, we introduce $e_i$ as the combination of the longitudinal formation error and the lateral formation error: $e_i = \sqrt{(e^s_i)^2 + (e^r_i)^2}$.

The first scenario involves three vehicles in a triangle-shaped formation (Fig.~\ref{fig:3v}) which can be used, for instance, for a snowplowing application~\cite{Ono2015}. The desired formation remains static during the entire simulation, while vehicles must avoid on-road obstacles and cross narrow corridors.   Fig.~\ref{fig:exp1traj} shows the computed trajectories of three vehicles using our framework. We remark that vehicles are able to quickly form the desired formation, and maintain it in the absence of obstacles. In the vicinity of obstacles (at the $100$-meters mark), vehicles are able to swerve around them and even temporarily deform the formation to pass. Fig.~\ref{fig:t1} presents the speed of each vehicle during the simulation. We observe that vehicle~2 decelerates at $t= \unit[13]{s}$ to avoid vehicle~1 when crossing a narrow corridor, showing the effectiveness of the proposed intra-formation collision avoidance strategy. Fig.~\ref{fig:err1} shows the formation errors of vehicle~1 and vehicle~2; we confirm that error converges quickly to 0 when the road is clear. The computation time profiles in Fig.~\ref{fig:comp1} demonstrates the real-time ability of the proposed algorithm.

\begin{figure*}[t!] 
\centering
\includegraphics[width=\linewidth]{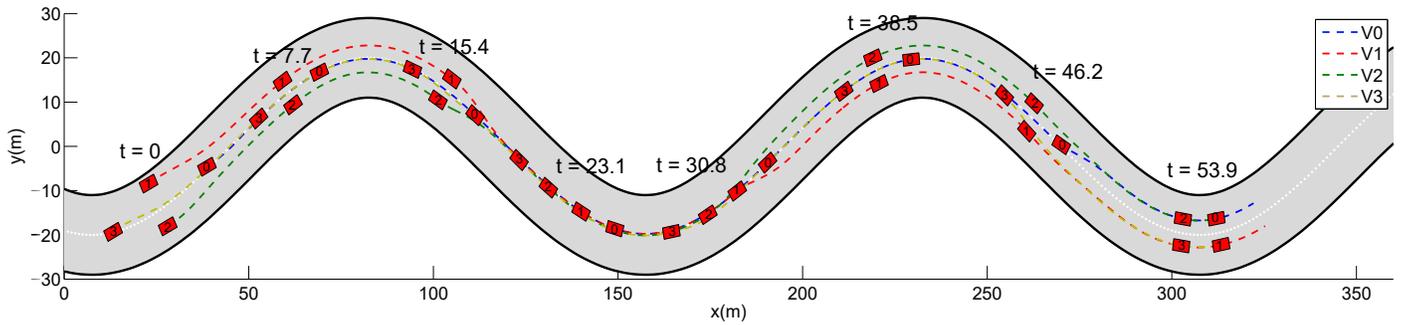}
\caption{Second scenario: trajectories of four vehicles.}
\label{fig:exp2traj}
\end{figure*}

The second scenario considers isomorphic formation changes for four vehicles on a curvy road. The formation sequence is illustrated in Example~\ref{ex3}. The time instants when we switch the formation are respectively $t=\unit[15.4]{s}$, $t=\unit[30.8]{s}$ and $t=\unit[46.5]{s}$. Fig.~\ref{fig:exp2traj} presents the trajectories of the four vehicles during the experiment. This experiment demonstrates that isomorphic formation changes can be performed smoothly while avoiding collisions. Moreover, the curvy nature of the road is handled nicely by our framework.

Videos of the two experiments are available on-line\footnote{ https://youtu.be/QIGIgCmBr0A}. 

\section{Conclusion}
\label{section:conclusion}

We have presented a distributed MPC-based formation control framework for car-like autonomous vehicles on the road. The proposed approach is distributed so that each vehicle locally computes a trajectory compatible with the formation control, using information exchanged through communication. We make use of logical rules to handle intra-formation collision avoidance. Moreover, we design a strategy for reconfiguration between isomorphic formations in a safe and timely manner. High-fidelity computer simulations have demonstrated the effectiveness of the approach in handling challenging on-road conditions with obstacles, road diets and curvy segments. 

This preliminary work clearly opens several research perspectives. A first avenue is to study and enhance the set of collision avoidance rules; here the rules~(\ref{eq:posde_cons}) have been chosen as they are symmetrical and intuitive. However, more complex rules can be considered. A second avenue of research is related to reconfiguration: the first simulation shows that we can avoid obstacles without needing an explicit reconfiguration. However, it seems rational to engage in a formal reconfiguration, which has been shown to be easily carried out in the second simulation. We will study the high-level decision-making process for reconfigurations. Finally, future work also includes the evaluation of this approach under realistic perception and communication conditions. This work is scheduled to be implemented in real vehicles under the framework of the European project AutoNET2030~\cite{autoNet-vision}.

\bibliographystyle{ieeetr}
\bibliography{biblio}

\end{document}